\documentclass[]{article}
\usepackage{proceed2e}
\usepackage[margin=1in]{geometry}

\usepackage[labelfont=bf]{caption}
\usepackage[mathscr]{eucal}
\usepackage[noend]{algpseudocode}
\usepackage[round]{natbib}
\usepackage[subrefformat=parens]{subcaption}
\usepackage[T1]{fontenc}
\usepackage[toc,page]{appendix}
\usepackage[usenames,dvipsnames,svgnames]{xcolor}
\usepackage{amsmath} 
\usepackage{amssymb}
\usepackage{mathtools}
\usepackage{listings}
\usepackage{algorithmicx}
\usepackage{algorithm}
\usepackage{amsfonts}
\usepackage{amsmath}
\usepackage{amssymb}
\usepackage{amsthm}
\usepackage{array}
\usepackage{bm}
\usepackage{booktabs}
\usepackage{colortbl}
\usepackage{enumitem}
\usepackage{fancyvrb}
\usepackage{graphicx}
\usepackage{lipsum}
\usepackage{listings}
\usepackage{mdframed}
\usepackage{multirow}
\usepackage{placeins}
\usepackage{standalone}
\usepackage{tabularx}
\usepackage{textcomp}
\usepackage{tikz}
\usepackage{url}
\usepackage{varwidth}
\usepackage{framed}

\setcitestyle{square}

\usepackage{times}
\newcommand{\prog}{\mathcal{P}}

\title{Probabilistic programs for inferring the goals of autonomous agents}

\author{
        {\bf Marco F.~Cusumano-Towner}$^1$, 
        {\bf Alexey Radul}$^1$,
        {\bf David Wingate}$^2$,
        {\bf Vikash K.~Mansinghka}$^1$\\
        $^1$Probabilistic Computing Project, Massachusetts Institute of Technology\\
        $^2$Computer Science Department, Brigham Young University
}

\begin{document}

\maketitle

\begin{abstract}
Intelligent systems sometimes need to infer the probable goals of people, cars, and robots, based on partial observations of their motion. This paper introduces a class of probabilistic programs for formulating and solving these problems. The formulation uses randomized path planning algorithms as the basis for probabilistic models of the process by which autonomous agents plan to achieve their goals. Because these path planning algorithms do not have tractable likelihood functions, new inference algorithms are needed. This paper proposes two Monte Carlo techniques for these ``likelihood-free'' models, one of which can use likelihood estimates from neural networks to accelerate inference. The paper demonstrates efficacy on three simple examples, each using under 50 lines of probabilistic code.
\end{abstract}
\vspace{-2mm}

\section{INTRODUCTION}
\vspace{-2mm}

Intelligent systems sometimes need to infer the probable goals of people, cars, and robots, based on partial observations of their motion. These problems are central to autonomous driving and driver assistance \citep{franke1998autonomous, urmson2008autonomous, aufrere2003perception}, but also arise in aerial robotics, reconnaissance, and security applications \citep{kumar2012opportunities, liao2006location, tran2008event}. In these settings, knowledge of the beliefs and goals of an agent makes it possible to infer their probable future actions.

Because the mental state of another agent is inherently unobservable and uncertain, it is natural to take a Bayesian approach to inferring it. Probabilistic models can be used to describe how an agent's latent high-level goals and beliefs about the environment interact to yield its probable actions. Most existing work along these lines has focused on modeling goal-directed behavior using Markov decision processes and related approaches from stochastic control \citep{baker2007goal, ziebart2009planning}. While promising, these approaches involve significant task-specific engineering. They also calculate policies that prescribe actions for every possible state of the world, sometimes in the inner loop of an inference algorithm. This leads to fundamental scaling challenges, even for simple environments and goal priors.

This paper introduces a class of probabilistic programs that formulate goal inference problems as approximate inference in generative models of goal-directed behavior. The proposed approach reflects three contributions:
First, agents are assumed to follow paths generated by fast randomized path planning code that can incorporate heuristics drawn from video game engines and robotics. This can scale to larger environments than approaches based on optimal control.
Second, hierarchical models for goals and paths are represented as probabilistic programs. This allows one to formulate a broad class of single- and multi-agent problems with common modeling and inference machinery. Ordinary probabilistic programming constructs can handle complex maps, hierarchical goal priors, and partially observed environments.
Third, this paper proposes an approach to real-time approximate inference, using neural networks to learn proposals for the internal choices made by any path planners. 
Together, these contributions lead to a practical proposal for goal inference that has the potential to scale to a broad class of real-world problems and real-time applications. We demonstrate the efficacy of prototype implementations of these algorithms on three simple examples, each written in under 50 lines of probabilistic code.

\begin{figure*}[t]
    \centering
    \includegraphics[width=1.00\textwidth]{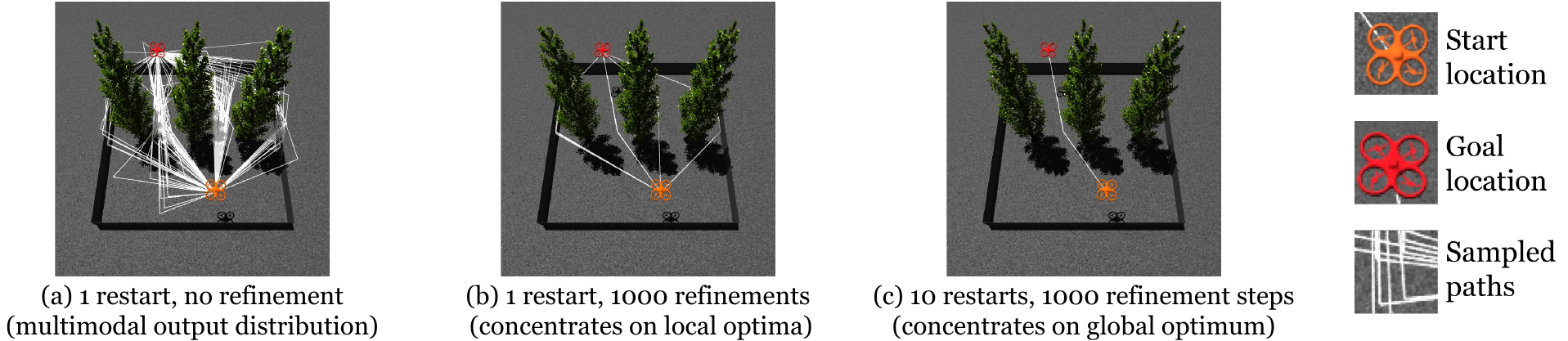}
    \caption{Each image shows paths from
    60 independent runs of \textproc{agent-path} for different settings of parameters $N$ (number of local refinement iterations) and $R$ (number of global restarts, among which the shortest path is selected).}
    \label{fig:planner}
\end{figure*}

Note that this proposal does not require planning algorithms to be rewritten as probabilistic programs, but instead allows optimized, low-level, or legacy planning codes to be treated as black boxes. This avoids the implementation and performance cost of rewriting an existing path planner in a high-level probabilistic programming language, and exposing the thousands of random choices it might make to generic inference algorithms. One difficulty is that such optimized black-box planners may well make too many internal random choices to have tractable input-output likelihoods. This paper proposes two novel Monte Carlo techniques for these ``likelihood-free'' models, each extending Metropolis-Hastings: (i) a {\em cascading resimulation} algorithm that makes joint proposals to ensure cancellation of the unknown likelihoods, and (ii) a {\em nested inference} algorithm that uses estimated likelihoods derived from inference over the internal random choices of the planner. Cascading resimulation is simple to implement, but nested inference enables use of a broad class of Monte Carlo, variational, and neural network mechanisms to handle the intractable likelihoods. 


\section{MODELING GOAL-DIRECTED BEHAVIOR USING RANDOMIZED PATH PLANNERS} \label{sec:SP}

This paper defines probabilistic models of goal-directed behavior using randomized path-planning algorithms. Algorithm~\ref{alg:planner} describes one such planner, called \textproc{agent-path}.  This planner can be applied to a broad class of environments with complex obstacles. The planner assumes a bounded 
two-dimensional space (e.g., the square $[0,1]^2$) and a world map $M$ that
is a set of polygonal obstacles. The planner takes
as input a start location $s \in [0, 1]^2$, a goal location $g \in
[0, 1]^2$, the map $M$, and a sequence of $T$ time points
$\mathbf{t} = (t_1, \ldots, t_T)$, and returns either a sequence of locations $\mathbf{z} \in
[0, 1]^{2T}$ on a path from $s$ to $g$ at each time $t_i$, or `no-path-found'.
The planner operates by growing a rapidly-exploring random tree (RRT) \citep{lavalle1998rapidly} from the start location $s$ to fill the space, searching for a clear line of sight between the tree and the goal. If a path is found, it is then refined to minimize its length using local optimization. Finally, the agent walks the path at a constant speed, producing the output locations $\mathbf{z}$. See Appendix~\ref{sec:planner-details} for more details.

\begin{figure*}[ht]
    \centering
    \includegraphics[width=1.00\textwidth]{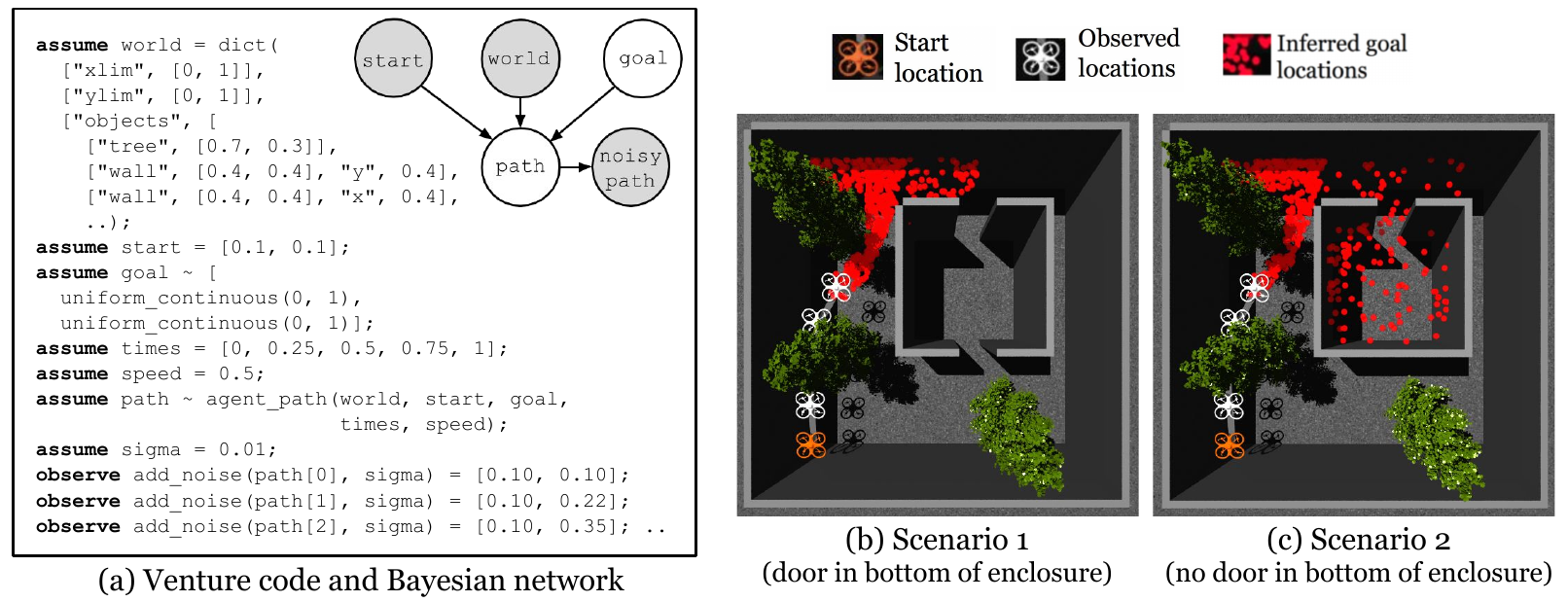}
    \caption{Inferring a simulated drone's goal from observed motion. (a) shows
a Venture model of a single drone that begins at
location \texttt{start} and moves to \texttt{goal}. \texttt{agent\_path}
is a likelihood-free path planner that models
the drone's goal-directed behavior using Algorithm~\ref{alg:planner}. (a) also shows a corresponding Bayesian network,
with observed nodes shaded. (b) and (c) show results of goal inference in this model
for two different environments, given the same observed path, using Cascading Resimulation Metropolis-Hastings. 
In Scenario 1, the drone's goal is likely outside the enclosure, since it did not go directly into the enclosure through the bottom. In Scenario 2, with bottom access to the enclosure blocked, a goal inside the enclosure is somewhat probable, as the drone's path no longer seems indirect.}
\label{fig:goal_inference}
\end{figure*}

Many variations of this planner are possible, including versions that take into account costs other than path length, and spaces encoding configurations other than geographic position (e.g., configuration spaces of an articulated robot). The planner parameters $N$ and $R$ trade off the cost of planning with the (probable) optimality of the paths (see Figure~\ref{fig:planner}). Figure~\ref{fig:goal_inference} and Figure~\ref{fig:fig_real_data} show this planner being used as a modeling primitive in the Venture probabilistic programming platform \citep{mansinghka2014venture}. The planner was implemented in C and imported as a foreign modeling primitive into Venture.
Venture supports likelihood-free primitives and design of custom inference strategies, including those of Section~\ref{sec:lfbn}.

\begin{algorithm}[H]
\footnotesize
    \begin{algorithmic}[1]
    \Require $\left\{
        \begin{array}{l}
            \mbox{World map}\; M;\; \mbox{Start, goal}\; s,g \in [0,1]^2\\
            \mbox{Time points}\; \mathbf{t} \in \mathbf{R}_{+}^{T}\\
            \mbox{Refinement amount}\; N;\; \mbox{Restarts}\; R\\
            \mbox{Max. \# tree nodes}\; J; \mbox{Min. \# tree nodes}\; S
        \end{array}
    \right.$
    \Procedure{rrt}{$M$, $s$, $g$}
    \State $V \gets \{s\}$ \Comment{Initialize tree with start $s$}
        \For{$j \gets 1$ to $J$} \Comment{$J$ tree growth iterations}
            \State $a \sim \mbox{Uniform}([0,1]\times[0,1])$ \Comment{Random point}
            \If{$M.\textproc{valid-state}(a)$}
                \State $b \gets \textproc{nearest-vertex}(V, a)$
                \State $\epsilon \sim \mbox{Uniform}([0, 1])$ 
                \State $c \gets \epsilon a + (1-\epsilon) b$ \Comment{Propose new vertex}
                \If{$M.\textproc{clear-line}(b, c)$}
                    \State $V.\textproc{add-edge}(b \to c)$ \Comment{Extend tree}
                    \If{$M.\textproc{clear-line}(c, g) \land j > S$}
                        \State $V.\textproc{add-edge}(c \to g)$
                        \State $\mathbf{return}\; \textproc{path-in-tree}(V, s, g)$
                    \EndIf
                \EndIf
            \EndIf
        \EndFor
        \State $\mathbf{return}\;$ `no-path-found'
    \EndProcedure
    \Procedure{plan-path}{$M$, $s$, $g$; $R$, $N$}
        \For{$r \gets 1$ to $R$} \Comment{Generate $R$ paths}
            \State $\mathbf{p}^{(r)} \sim \textproc{rrt}(M, s, g)$
            \State $\mathbf{p}^{(r)} \gets \textproc{simplify-path}(\mathbf{p}^{(r)})$
            \State $\mathbf{p}^{(r)} \sim \textproc{refine-path}(M, s, g, \mathbf{p}^{(r)})$
            \State $d^{(r)} \gets \textproc{path-length}(\mathbf{p}^{(r)}, s, g)$
        \EndFor
        \State $r^* \gets \textproc{argmin}(\mathbf{d})$ \Comment{Select best of $R$ paths}
        \State $\mathbf{return}\; \mathbf{p}^{(r^*)}$
    \EndProcedure
    \Procedure{agent-path}{$M$, $s$, $g$, $\mathbf{t}$; $R$, $N$}
        \State $\mathbf{p} \sim \textproc{plan-path}(M, s, g; R, N)$ \Comment{Abstract path}
        \State $\mathbf{z} \gets \textproc{walk-path}(\mathbf{p}, \mathbf{t})$ \Comment{Locations at times $\mathbf{t}$}
        \State $\mathbf{return}\; \mathbf{z}$
    \EndProcedure
    \end{algorithmic}
    \caption{Model of an agent's path given destination}
    \label{alg:planner}
\end{algorithm}

\section{INFERENCE IN PROBABILISTIC PROGRAMS WITH LIKELIHOOD-FREE PRIMITIVES} \label{sec:lfbn}
\vspace{-2mm}
The path planner \textproc{agent-path} of Algorithm~\ref{alg:planner} can be used in a probabilistic program either by implementing the planner in a probabilistic programming language, or by treating the planner as a primitive random choice. We treat the planner as a random choice, as this allows use of an optimized C implementation of the planner.
However, probabilistic programming languages such as Church, Stan, BLOG, and Figaro all require random choices to have tractable marginal likelihoods \citep{goodman2012church, carpenter2016stan, milch20071, pfeffer2009figaro}. 
Computing the marginal likelihood of \textproc{agent-path} for outputs $\mathbf{z}$ and inputs $M$, $s$, $g$, and $\mathbf{t}$ would involve an intractable intregral over the (thousands of) internal random choices made in \textproc{agent-path}.

This section introduces two Monte Carlo strategies for inference in probabilistic programs that include random choices with intractable marginal likelihoods, referred to as ``likelihood-free'' primitives. The first strategy, shown in Algorithm~\ref{alg:cascading-mh}, is called {\em Cascading Resimulation Metropolis-Hastings}; it makes block proposals to likelihood-free random choices, exploiting cancellation of the unknown likelihoods. The second, shown in Algorithm~\ref{alg:nested-inference-mh}, is called {\em Nested Inference Metropolis-Hastings}; it uses Monte Carlo estimates of the unknown likelihoods in place of the likelihoods themselves.
Although simple techniques like likelihood-weighting can also be used in the presence of likelihood-free primitives, they tend to work well only when a global proposal that is well-matched to the posterior is available.
The algorithms we introduce do not have this limitation.

We first introduce notation. Let $\mathcal{T}$ be the set of primitive random choices available to
a probabilistic program (e.g. $\{\textproc{flip}, \textproc{uniform\_continuous}, \textproc{agent-path}\}$). 
For $t \in \mathcal{T}$, let $\mathcal{X}_{t}$ denote the set of valid arguments for the primitive, let 
$\mathcal{Z}_t$ denote the set of possible outputs, and let $p_{t}(z; x)$ denote the 
marginal likelihood of output $z \in \mathcal{Z}_t$ given arguments
$x \in \mathcal{X}_t$, where $\int p_t(z; x) dz = 1$ for all $x \in \mathcal{X}_t$. 
We do not require evaluation of $p_{t}(z; x)$ to be computationally tractable.

Following \citet{wingate2011lightweight}, for a probabilistic program $\prog$, 
we assume there is a name $i \in \mathcal{I}$ assigned to
every possible random choice, for some countable $\mathcal{I}$. We assume
that distinct random choices are assigned unique names within every
execution of $\prog$. The set of names used in an execution is some finite set
$I \subseteq \mathcal{I}$.  We require that all
random choices with name $i$ are of the same type $t_i \in \mathcal{T}$.
Each unique completed execution of $\prog$
can therefore be represented as the finite set of names $I \subseteq \mathcal{I}$
of those random choices made in the execution, together with the result values.  We denote these results $z \in \bigtimes_{i \in I} \mathcal{Z}_{t_i}$,
and denote this complete package $\rho = (I, z)$. The tuple $\rho$ is called an \emph{execution trace} of the probabilistic program $\prog$.

This paper focuses on probabilistic
programs where $I$ is the same for all executions --- that is, the set of random choices made is not affected by any of those choices. Relaxations of this are left for future work; more general formalizations of probabilistic programs can be found
in \citep{wingate2011lightweight, mansinghka2014venture}. 

We consider random choice $j$ to \emph{depend on} random choice $i$
if changing the result $z_i$ of $i$ can lead to a change in the
inputs $x_j$ of $j$, even if all other results $z_{I\setminus\{i\}}$ are held fixed.
We assume that it is possible to construct a directed acyclic dependency
graph $G = (I, E)$ among random choices $I$, where an edge $(i, j) \in E
\subset I \times I$ exists if and only if
random choice $j$ depends on random choice $i$ in the above sense. The parents of a random choice $j$ are denoted by $\pi_G(j) := \{i \in I: (i, j) \in E\}$. The
arguments $x_j$ of each random choice $j$ are then a (deterministic) function $f_j$ of the results of
random choices in $\pi_G(j)$, which are denoted $z_{\pi_G(j)}$; we write $x_j =
f_j(\pi_G(j))$. Let $c_G(i) := \{ j \in I : (i, j) \in E\}$ denote the
`children' of choice $i$. Also, let $F \subseteq I$ denote the random choices with intractable likelihoods (the ``likelihood-free'' choices). Let $C \subseteq (I \setminus F)$
denote the random choices that are constrained based on data, which must have tractable likelihoods. Let $z_C \in \bigtimes_{i \in C}
\mathcal{Z}_i$ denote the values we are constraining those random choices to.
The joint probability density of an
execution trace $\rho = (I, z)$ is:
\begin{equation}
p(z) := \prod_{i \in I} p_{t_i}(z_i; f_i(z_{\pi_G(i)}))
\end{equation}
where we have omitted the dependence on $I$ because it is the same for all
executions. 

\subsection{CASCADING RESIMULATION METROPOLIS-HASTINGS}


How can a probabilistic program cope with complex, likelihood-free primitives?  Our core insight is that if the proposal distribution $m(z'_i;\cdot)$ for a random choice $z_i$ is equal to the prior $p_{t_i}(z_i'; f_i(z_{\pi_G(i)}))$, then the likelihoods will cancel in a Metropolis-Hastings (MH) acceptance ratio and therefore do not need to be explicitly computed. 
Sampling from the prior is achieved simply by simulating the random choice.
A (prototypical) acceptance ratio looks like this:
\[
\arraycolsep=0.1pt
\begin{array}{ccccccccc}
\alpha = \;&\frac{ p_{t_i}(z'_i; \cdot) }{p_{t_i}(z_i; \cdot)} & \cdot & \frac{m(z_i;\cdot)}{m(z'_i;\cdot)} & \;=\; & \;\frac{ p_{t_i}(z'_i; \cdot) }{p_{t_i}(z_i; \cdot)}&\cdot&
\frac{ p_{t_i}(z_i; \cdot) }{p_{t_i}(z'_i; \cdot)}&
\;= 1\\
& \mbox{{\small target }} &&  \mbox{{\small proposal }} && \mbox{{\small target }} &&  \mbox{{\small proposal }} &
\end{array}
\]

We use blocked proposals in which a change to a likelihood-free choice is proposed from the prior whenever a proposal is made to any of its parents. 
A likelihood-free choice that is proposed may itself have likelihood-free choices as children, in which case these children are also proposed, generating a cascade of proposals.
Algorithm~\ref{alg:cascading-mh} shows the Cascading Resimulation MH transition operator, which extends an initial custom proposal $m(z_i'; z)$ to random choice $i$ (which must not be likelihood-free) to also include any likelihood-free random choices $H$ in the cascade, such that the intractable likelihoods cancel.

\begin{algorithm}[h]
\footnotesize
\begin{algorithmic}[1] 
\Require $\left\{\begin{array}{l} 
\mbox{Prob. program with dep. graph}\; G = (I,E)\\
\mbox{Likelihood-free random choices}\; F \subseteq I\\
\mbox{Proposed-to random-choice}\; i \in (I \setminus F)\\
\mbox{Custom proposal density}\; m(z_i'; z)\; \mbox{for choice $i$}\\
\mbox{Previous values $z$ for all random choices}\\
\end{array} \right.$
    \State $z_{i}' \sim m(\cdot; z)$ \Comment{Propose a new value for choice $i$}
    \State $z_{I \setminus \{i\}}' \gets z_{I \setminus \{i\}}$ \Comment{Initially, no change to other choices}
    \State $\ell \gets 1$ \Comment{Unnormalized target density for previous values}
    \State $\ell' \gets 1$ \Comment{Unnormalized target density for proposed values}
    \State $B \gets \{i\} \cup c_G(i)$ \Comment{Ask for likelihoods from $j \in B$}
    \State $H \gets \{\}$ \Comment{Likelihood-free cascade participants}
    \State $A \gets \{i\}$ \Comment{Visited choices with tractable likelihoods}
    \While{$|B| > 0$}
        \State $j \gets \textproc{pop}(B)$ \Comment{Pop in topological order}
        \If{$j \in F$} \Comment{Choice $j$ is likelihood-free}
            \State $z_j' \sim p_{t_j}(\cdot; f_j(z_{\pi_G(j)}'))$ \Comment{Propose from prior}
            \State $\textproc{insert}(B, c_G(j))$ \Comment{Ask for child likelihoods}
            \State $H \gets H \cup \{j\}$
        \Else \Comment{Choice $j$ has tractable likelihood}
            \State $\ell \gets \ell \cdot p_{t_j}(z_j; f_j(z_{\pi_G(j)}))$
            \State $\ell' \gets \ell' \cdot p_{t_j}(z_j'; f_j(z_{\pi_G(j)}'))$
            \State $A \gets A \cup \{j\}$
        \EndIf
    \EndWhile
    \State $\alpha \gets (\ell' / \ell) \cdot (m(z_i; z') / m(z_i'; z))$ \Comment{MH ratio}
    \State $s \sim \mbox{Uniform}(0, 1)$
    \If{$s \le \alpha$}
        \State $z \gets z'$ \Comment{Accept}
    \EndIf
\end{algorithmic} 
\caption{Single-site Cascading Resimulation Metropolis-Hastings transition}
\label{alg:cascading-mh} \end{algorithm}

Algorithm~\ref{alg:cascading-mh} is a Metropolis-Hastings transition over the random choices $\{i\} \cup H$ with target density equal to the local posterior $p\left(z_{\{i\} \cup H} | z_{I \setminus \{\{i\} \cup H\}}\right)$,
and with proposal density:
\begin{equation}
    m(z_i'; z) \prod_{j \in H} p_{t_j}(z_j'; f_j(z_{\pi_G(j)}'))
\end{equation}
The Metropolis-Hastings acceptance ratio is:
\begin{align}
\alpha &= \left( \frac{\prod_{j \in H \cup A} p_{t_j}(z_j'; f_j(z_{\pi_G(j)}'))}
{\prod_{j \in H \cup A} p_{t_j}(z_j; f_j(z_{\pi_G(j)}))}\right. \notag \\
&\qquad\cdot
\left.\frac{m(z_i; z') \prod_{j \in H} p_{t_j}(z_j; f_j(z_{\pi_G(j)}))}
{m(z_i'; z) \prod_{j \in H} p_{t_j}(z_j'; f_j(z_{\pi_G(j)}'))}\right) \notag \\
&= \frac{m(z_i; z') \prod_{j \in A} p_{t_j}(z_j'; f_j(z_{\pi_G(j)}'))}
{m(z_i'; z) \prod_{j \in A} p_{t_j}(z_j; f_j(z_{\pi_G(j)}))}
\end{align}
We illustrate Cascading Resimulation MH in Figure~\ref{fig:goal_inference}, on the task of inferring the goal of a simulated drone in
an observed environment.

\subsection{NESTED INFERENCE METROPOLIS-HASTINGS} \label{sec:nested-inference}
In some problems, Cascading Resimulation MH will generate many expensive simulations of likelihood-free choices, most of which will be rejected.
For these problems, and for real-time applications, we propose an alternative Metropolis-Hastings algorithm, called Nested Inference MH, that uses Monte Carlo estimates of the intractable likelihoods in the acceptance ratio. 
The likelihood estimates are obtained using auxiliary ``nested inference'' algorithms, which sample probable values for the internal random choices made by a likelihood-free choice (e.g. a randomized planning algorithm) given its inputs and outputs, and calculate a weight that can be used to form an importance sampling estimate of the unknown likelihood.

Nested Inference MH is based on an interpretation of likelihood-free random choices like \textproc{agent-path} as probabilistic programs in their own right. 
Let $u \in \mathcal{U}_t$ be an execution trace of a likelihood-free random choice of type $t$. 
We denote the joint density on execution traces $u$ and return values $z$ of the random choice, given input arguments $x$, by $p_t(u, z; x)$. 
The marginal likelihood of the random choice is given by the (intractable)
integral $p_t(z; x) = \int p_t(u, z; x) du$. 
We denote the conditional trace density for arguments $x$ and output $z$ by $p_t(u | z; x) := p_t(u, z; x)/p_t(z;x)$.

Nested inference assumes the existence of a {\em nested inference algorithm} that samples execution traces $u$ according to some density $q_t(u; x, z)$ that approximates the conditional density on traces of the likelihood-free choice, i.e., $q_t(u; x, z) \approx p_t(u | z; x)$. We require that $q_t(u; x, z) > 0$ for all $u$ where $p_t(u | z; x) > 0$. Using the nested inference algorithm as an importance sampler, we produce an unbiased importance sampling estimate $\hat{p}_t(z;x)$ of the random choice's intractable likelihood for arguments $x$ and output $z$ by sampling $K$ times $u_k \sim q_t(\cdot; x, z)$ from the inference algorithm, as follows:
\begin{equation}
\hat{p}_t(z;x) := \frac{1}{K} \sum_{k=1}^K \frac{p_t(u_k, z; x)}{q_t(u_k; x, z)}\; \mbox{for} \;u_k \sim q_t(\cdot; x, z).
\end{equation}

Nested inference also assumes that the ratio $p_t(u, z; x) / q_t(u; x, z)$ can be evaluated. 
While in principle the nested inference algorithm can be produced by recoding the likelihood-free primitive in a high-level probabilistic programming language, this is by no means required, nor do we expect it to be the common case.
In this paper, we focus on nested inference algorithms that use learned neural networks.

The accuracy of the likelihood estimate is determined by the accuracy of the nested inference algorithm. Specifically, for $K = 1$ the variance of the estimate is:
\begin{align}
    &\mbox{Var}_{u \sim q(\cdot; x, z)} \left[ \frac{p_t(u, z; x)}{q_t(u;x,z)} \right] \notag \\
    &\qquad\propto \mbox{D}_{\chi^2}\left( p_t(u | z; x) || q_t(u; x, z) \right),
\end{align}
where $\mbox{D}_{\chi^2}$ denotes the chi-square divergence \citep{nielsen2014chi}, and where $p_t(u|z;x)$ and $q_t(u;x,z)$ on the right-hand side represent density functions over $u$, not specific density values. Similarly, we can view $\log (p_t(u, z; x) / q_t(u; x, z))$ for $u \sim q(\cdot; x, z)$ as a (biased) estimator of $\log p_t(z;x)$, where the bias is:
\begin{align}
    &\mbox{E}_{u \sim q_t(\cdot; x, z)} \left[ \log \frac{p_t(u, z; x)}{q_t(u;x,z)} \right] - \log p_t(z;x) \notag \\
    &\qquad= -\mbox{D}_{KL}( q_t(u; x, z) || p_t(u | z; x) ),
\end{align}
where $\mbox{D}_{KL}$ denotes the Kullback-Leibler (KL) divergence \citep{kullback1951information}.

\subsubsection{Nested Inference Metropolis-Hastings}
Algorithm~\ref{alg:nested-inference-mh} describes a Nested Inference MH transition in which a custom proposal is made to a likelihood-free random choice $i$ that uses estimated likelihoods produced using a nested inference algorithm. It assumes that all children of $i$ also have nested inference algorithms themselves. Heterogeneous configurations are also possible.

\begin{algorithm}[H] 
\footnotesize
\begin{algorithmic}[1] 
\Require $\left\{\begin{array}{l} 
\mbox{Prob. program with dep. graph}\; G = (I,E)\\
\mbox{Proposed-to random choice}\; i\\
\mbox{Custom proposal density}\; m(z_i'; z)\\
\mbox{Previous values $z$ for all random choices}\\
\mbox{Previous likelihood estimates $\ell$ for all choices}
\end{array} \right.$
    \State $z_{i}' \sim m(\cdot; z)$ \Comment{Propose a new value for choice $i$}
    \State $z_{I \setminus \{i\}}' \gets z_{I \setminus \{i\}}$ \Comment{No change to other choices}
    \For{$k \gets 1$ to $K$}
        \State $u_{i,k}' \sim q_{t_i}(\cdot; x_i, z_i')$ \Comment{Choice $i$ nested inference}
    \EndFor
    \State $\ell_i' \gets \frac{1}{K} \sum_{k=1}^K \frac{p_{t_i}(u_{i,k}', z_i'; x_i)}{q_{t_i}(u_{i,k}'; x_i, z_i')}$ \Comment{Estimate $p_{t_i}(z_i';x_i)$}
    \For{$j \in c_G(i)$}
        \For{$k \gets 1$ to $K$}
            \State $u_{j,k}' \sim q_{t_j}(\cdot; x_j', z_j)$ \Comment{Choice $j$ nested inference}
        \EndFor
        \State $\ell_j' \gets \frac{1}{K} \sum_{k=1}^K \frac{p_{t_j}(u_{j,k}', z_j; x_j')}{q_{t_j}(u_{j,k}'; x_j', z_j)}$ \Comment{Estimate $p_{t_j}(z_j;x_j')$}
    \EndFor
    \State $\alpha \gets \left( \prod_{j \in \{i\} \cup c_G(i)} \frac{\ell_j'}{\ell_j}\right) \cdot \left( \frac{m(z_i; z')}{m(z_{i}'; z)} \right)$
        \State $s \sim \mbox{Uniform}(0, 1)$
        \If{$s \le \alpha$}
            \State $z_i \gets z_i'$ \Comment{Accept}
            \For{$j \in \{i\} \cup c_G(i)$}
                \State $\ell_j \gets \ell_j'$\Comment{Update density estimates}
            \EndFor
        \EndIf 
\end{algorithmic} 
\caption{Single-site Nested Inference Metropolis-Hastings transition}
\label{alg:nested-inference-mh}
\end{algorithm}

Although this transition uses Monte Carlo estimates of likelihoods in the acceptance ratio, it is a standard Metropolis-Hastings transition on
an extended state space that includes the result $z_i$ of the proposed-to random choice $i$, $K$ traces $u_{i,k}$ of the proposed-to random choice, and $K$ traces $u_{j,k}$ of each child $j$ of the proposed-to random choice.
The target density on the extended space is:
\begin{equation}
\resizebox{\hsize}{!}{
    $\displaystyle p(z_i | z_{I \setminus \{i\}}) \prod_{j \in \{i\} \cup c_G(i)} \frac{1}{K} \sum_{k=1}^K p_{t_j}(u_{j,k} | z_j; x_j) \prod_{\substack{r=1\\r \ne k}}^K q_{t_j}(u_{j,r}; x_j,z_j)$
}
\end{equation}
The proposal density on the extended space is:
\begin{equation}
    m(z_i'; z) \prod_{j \in \{i\} \cup c_G(i)} \prod_{k=1}^K q_{t_j}(u_{j,k}'; x_j, z_j)
\end{equation}
The values $z_j$ of other random choices $j \not \in \{i\} \cup c_G(i)$ are constant.
See Appendix~\ref{sec:nested-inference-derivations} for derivation.
The marginal density of $z_i$ in the extended target density is the local posterior $p(z_i | z_{I \setminus \{i\}})$ for the result of random choice $i$ given the values of all other random choices.
Single-site Nested Inference MH transitions that propose to different random choices $i$ but 
use the same database of nested-inference likelihood estimates $\ell$ can be composed to form Markov chains that converge to the posterior $p(z_{I \setminus C} | z_C)$.

Our use of unbiased likelihood estimates in place of the true 
likelihoods when computing the Metropolis-Hastings
acceptance ratio in Algorithm~\ref{alg:nested-inference-mh} is closely related to pseudo-marginal MCMC \citep{andrieu2009pseudo} and particle MCMC \citep{andrieu2010particle}. 
Indeed, each single-site Nested Inference MH transition can be seen as a compositional variant of a `grouped independence MH' transition \citep{beaumont2003estimation} in which several pseudo-marginal likelihoods (one for each random choice $j \in \{i\} \cup c_G(i)$) are used in the same update. The database of nested-inference likelihood estimates $\ell$ stores the `recycled' pseudo-marginal likelihood estimates from previous transitions. 

The convergence rate of a Markov chain based on Nested Inference MH transition operators depends on the accuracy of the nested inference algorithm and $K$. In the limit of exact nested inference algorithm ($q_t(u;x,z) = p_t(u|z;x)$) the likelihood estimates are exact, and the algorithm is identical to standard Metropolis-Hastings. If the nested inference algorithm is very inaccurate, it may routinely propose traces $u$ that are incompatible with the output $z$ of the random choice, resulting in low acceptance rates. Better characterizing how the convergence rate depends on the accuracy of the nested inference algorithms and on $K$ is an important area for future work.

\subsubsection{Learning a nested inference algorithm} \label{sec:learning}
It is possible to learn a nested inference algorithm $q_t(u; x, z)$ that approximates $p_t(u|z;x)$.
The idea of learned inference for probabilistic generative models goes back at least to \citet{morris2001recognition} and has also been used in \citet{stuhlmuller2013learning} and \citet{kingma2013auto}. We apply this idea to nested inference
as follows. Let $q_{t,\theta}(u;x,z)$ denote a nested inference algorithm that is parameterized by $\theta$ --- for example, $\theta$ might be the weights of a neural network used as part of the inference algorithm. We establish a training distribution $d_t(x)$ over the arguments to the primitive $t$, and approximately solve the following optimization problem:
\begin{align*}
\min_{\theta} \left\{ \mbox{E}_{\begin{subarray}{l}x \sim d_t(\cdot)\\z|x \sim p_t(\cdot; x)\end{subarray}} \left[ \mbox{D}_{KL}(p_t(u | z; x) || q_{t, \theta}(u; x, z)) \right]
\right\}\\
= \min_{\theta} \left\{\mbox{E}_{\begin{subarray}{l}x \sim d_t(\cdot)\\z|x \sim p_t(\cdot; x)\\u|x,z \sim p_t(\cdot | z; x)\end{subarray}} \left[ \log \frac{p_t(u | z; x)}{q_{t, \theta}(u; x, z)} \right]
\right\}
\end{align*}
The goal is for $q_{t,\theta}(u;x,z)$ to approximate $p_t(u|z;x)$ well (i.e., have small KL divergence) for typical input arguments $x \sim d_t(\cdot)$. We approximate this objective function by drawing $M$ independent sets of input arguments $x^{(i)}$ from the training distribution, and running a traced execution of the likelihood-free random choice (e.g. planner) on each set of arguments, recording\footnote{This training regime cannot be applied to a true black-box
path planner, since a recording of its internal randomness is now necessary.  However,
such recordings can be produced from a straightforwardly instrumented version of the algorithm. The likelihood estimator for the planner can still be treated as a black-box by the Nested Inference MH transition.} the trace $u^{(i)}$ and  output $z^{(i)}$:
\[
\begin{array}{rl}
    x^{(i)} \sim d_t(\cdot)& \mbox{{Sample planner arguments}}\vspace{2mm}\\
    u^{(i)}, z^{(i)} \sim p_t(\cdot, \cdot; x^{(i)})& \hspace{-2mm} \begin{array}{l}
        \mbox{{Run likelihood-free planner,}}\\
        \mbox{{record trace $u^{(i)}$, output $z^{(i)}$}}
        \end{array}
\end{array}
\]
We use the resulting dataset $D = \{(x^{(i)}, z^{(i)}, u^{(i)}) : i = 1\dots M\}$ to define an approximate objective function $J_D(\theta)$ that is an unbiased estimate of the original objective function:
\begin{align}
J_D(\theta) &:= \frac{1}{M} \sum_{i=1}^M \log \frac{p_t(u^{(i)} | z^{(i)}, x^{(i)})}{q_{t,\theta}(u^{(i)}; x^{(i)}, z^{(i)})}\\
&= C - \frac{1}{M} \sum_{i=1}^M \log q_{t,\theta}(u^{(i)}; x^{(i)}, z^{(i)})
\end{align}
where $C$ does not depend on $\theta$. Note that minimizing $J_D(\theta)$ over $\theta$ is equivalent to maximizing the log-likelihood of the data $D$. Because we use forward simulations to produce $u^{(i)}, z^{(i)}$ jointly from $p_t(\cdot, \cdot; x^{(i)})$, we have one exact conditional sample $u^{(i)} | z^{(i)} \sim p_t(\cdot | z^{(i)}; x^{(i)})$ for each training example.

\section{EXAMPLE APPLICATIONS} \label{sec:applied}
We have implemented four example applications, designed to illustrate the flexibility of our framework:
\begin{enumerate}
    \item Inferring the probable goal of a simulated drone. This example shows that small changes to the environment, such as including an additional doorway, can yield large changes in the inferred goals.
    \item Inferring the probable goal of a simulated drone with a more complex planner.  Specifically, we model the drone as following a multi-part path produced by a planner that first chooses a waypoint uniformly at random and then recursively solves the two path planning problems induced by the choice of waypoint. This example shows (a) applicability of the framework to more complex models of goal-directed behavior, and (b) that Nested Inference MH with a learned neural network can outperform Cascading Resimulation MH.
    \item Inferring whether or not two people walking around tables in a room are headed for the same goal or different goals. This example demonstrates applicability to simple hierarchical models for goals and also demonstrates applicability to real-world (as opposed to synthetic) data.
    \item Jointly inferring a simulated agent's goals and its beliefs about an obstacle in the map whose location, size, and orientation is unknown to the probabilistic program. This example is described in the appendix due to space constraints.
\end{enumerate}

\begin{figure*}[ht]
    \centering
    \includegraphics[width=1.00\linewidth]{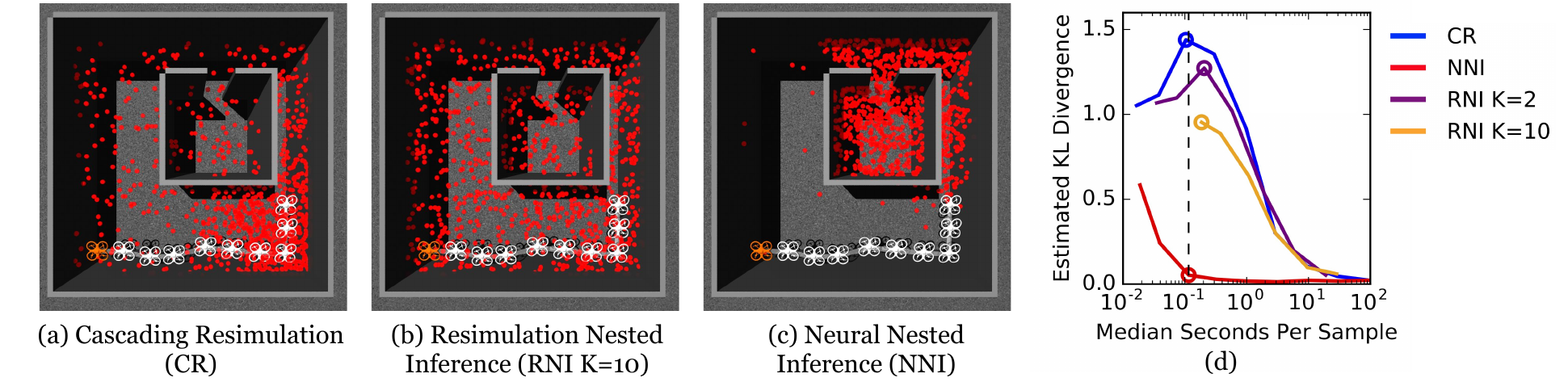}
    \caption{Comparison of three Metropolis-Hastings (MH) strategies for goal inference in a model that uses the \textproc{agent-waypoint-path} planner, which models an agent's motion using an unknown waypoint. (a), (b) and (c) show 960 independent approximate posterior goal samples (red) obtained using each strategy for similar run-times, given known map, start location (orange), and observations (white). Cascading Resimulation MH (CR) and Resimulation Nested Inference MH (RNI) do not give accurate inferences in real-time because they propose the waypoint from the prior. Neural Nested Inference MH (NNI) uses a neural network to propose the waypoint and gives accurate results in real-time (median 115 ms per sample). (d) shows estimated KL divergences from gold-standard samples to each of the strategies as the number of MH transitions are varied. Circles in (d) show the amount of computation used for (a,b,c). }
\label{fig:nested-combined}
\end{figure*}

\subsection{EXAMPLE 1: SENSITIVITY OF GOAL INFERENCE TO SMALL MAP CHANGES}

Figure~\ref{fig:goal_inference} shows a comparison of goal inference in two different maps given the same observations. The map for the scenario on the left has an enclosure with two openings, one on the top and one on the bottom, while the map for the scenario on the right has a single opening. In the map on the left, the inferred goal samples fall outside the enclosure, because if the drone intended to go inside the enclosure, it could have taken a much shorter path. In the map on the right, a significant fraction of goal samples fall inside the enclosure, as relatively efficient paths into the enclosure go through the partial trajectory that has been observed so far. Samples shown are the final states of 480 independent replicates of a Markov chain initialized from the prior, with $1000$ Cascading Resimulation MH transitions (Algorithm~\ref{alg:cascading-mh}) using the prior as the proposal. Planner parameters are $R = 10$, $N = 1000$, $\epsilon = 0.01$, $v = 0.5$, $J = 10000$, $S = 2000$.

\subsection{EXAMPLE 2: HANDLING PATH PLANNERS WITH WAYPOINTS VIA NESTED INFERENCE}
Next, we used a model where the agent may choose a waypoint and separately plan a path to the waypoint and a path from the waypoint to the goal (\textproc{agent-waypoint-path}, Algorithm~\ref{alg:agent-waypoint-path}). Unlike the simpler $\textproc{agent-path}$ model, which typically samples from a small number of modes concentrated at efficient routes from the start to the goal, \textproc{agent-waypoint-path} yields paths that are unpredictable without knowledge of the waypoint. Parameters $R$ and $N$ of \textproc{plan-path} are omitted for simplicity. We consider the same goal inference task as in Example 1 but with the alternative planner.
Cascading Resimulation MH performs poorly on this task, because the prior is a poor proposal for the internal random choices of \textproc{agent-waypoint-path}.
\begin{algorithm}[h]
\footnotesize
    \begin{algorithmic}[1]
    \Require $\left\{
        \begin{array}{l}
            \mbox{World map}\; M;\;\;\mbox{Start, goal}\; s,g \in [0,1]^2\\
            \mbox{Time points}\; \mathbf{t} \in \mathbf{R}^{T}_{+}
        \end{array}
    \right.$
    \Procedure{agent-waypoint-path}{$M$, $s$, $g$, $\mathbf{t}$}
        \State $g' \sim \mbox{Uniform}([0,1] \times [0,1])$ \Comment{Pick waypoint}
        \State $w \sim \mbox{Bernoulli}(0.5)$ \Comment{Use waypoint?}
        \If{$w$}
            \State $\mathbf{p}_1 \sim \textproc{plan-path}(M, s, g')$ \Comment{{\small Start-waypoint}}
            \State $\mathbf{p}_2 \sim \textproc{plan-path}(M, g', g)$ \Comment{{\small Waypoint-goal}}
            \State $\mathbf{p} = (\mathbf{p}_1, \mathbf{p}_2)$ \Comment{Concatenate paths}
        \Else
            \State $\mathbf{p} \sim \textproc{plan-path}(M, s, g)$ \Comment{Start to goal}
        \EndIf
        \State $\mathbf{\tilde{z}} \gets \textproc{walk-path}(\mathbf{p}, \mathbf{t})$ \Comment{Locations at times $\mathbf{t}$}
        \State $\mathbf{z} \sim \textproc{add-noise}(\mathbf{\tilde{z}})$ \Comment{Add noise to locations}
        \State $\mathbf{return}\; \mathbf{z}$ \Comment{Return noisy agent locations}
    \EndProcedure
    \end{algorithmic}
    \caption{Pseudo-code for a likelihood-free primitive that models observed motion of an agent with known goal but optional unknown waypoint}
    \label{alg:agent-waypoint-path}
\end{algorithm}
\vspace{-1mm}

Algorithm~\ref{alg:agent-waypoint-path-nested-inference} shows a nested inference algorithm for \textproc{agent-waypoint-path} that uses a neural network to propose the waypoint ($g'$) and whether the waypoint is used ($w$), given the goal and observations, and then executes the rest of the planner, conditioned on $w$ and $g'$.
The network was trained on 10,000 runs of \textproc{agent-waypoint-path} with random goal input $g \sim \mbox{Uniform}([0, 1]^2)$ and fixed world map $M$ and start $s$. The nested inference algorithm splits the trace $u$ of \textproc{agent-waypoint-path} into $u_1 = (w, g')$ and $u_2$ (the random choices made within executions of \textproc{plan-path}), so that $u = (u_1, u_2)$. The density of the nested inference algorithm is then $q_t(u; x, z) = q_{\theta}(u_1; x, z) p_t(u_2; x)$, and the density ratio $p_t(u, x, z) / q_t(u; x, z)$, which is used by Nested Inference MH when estimating the planner likelihoods, simplifies to $p_t(z | u; x) / q_{\theta}(u_1; x, z)$. To evaluate this ratio, we separately evaluate the density $p_t(z | u;, x)$ of \textproc{add-noise} and the density $q_{\theta}(u_1; x, z)$ of the neural network's stochastic outputs.
\begin{algorithm}[H]
\footnotesize
    \begin{algorithmic}[1]
    \Require $\left\{
        \begin{array}{l}
            \mbox{Arguments to planner}\; \mathbf{x} = (M, s, g, \mathbf{t})\\
            \mbox{Hypothetical output of planner}\; \mathbf{z}
        \end{array}
    \right.$
        \State $w, g' \sim q_{\theta}(\cdot; \mathbf{x}, \mathbf{z})$ \Comment{{\small Sample waypoint from neural net.}}
        \If{$w$}
            \State $\mathbf{p}_1 \sim \textproc{plan-path}(M, s, g')$ \Comment{Start to waypoint}
            \State $\mathbf{p}_2 \sim \textproc{plan-path}(M, g', g)$ \Comment{Waypoint to goal}
            \State $\mathbf{return}\; (\mathbf{p}_1, \mathbf{p}_2)$
        \Else
            \State $\mathbf{return}\; \textproc{plan-path}(M, s, g)$ \Comment{Start to goal}
        \EndIf
    \end{algorithmic}
    \caption{Using a neural network for nested inference in the $\textproc{agent-waypoint-path}$ path planner}
    \label{alg:agent-waypoint-path-nested-inference}
\end{algorithm}
We compared three strategies for goal inference: Nested Inference MH using Algorithm~\ref{alg:agent-waypoint-path-nested-inference} and $K=1$, Cascading Resimulation MH, and Nested Inference MH using a ``resimulation'' nested inference algorithm $(q_t(u;x,z) = p_t(u;x))$ and $K=2,10$. 
Figure~\ref{fig:nested-combined} shows that neural Nested Inference MH converges faster than the other strategies. Planner parameters were the same as in Example 1. All inference strategies were implemented using a custom Python inference library. Integration of Nested Inference MH with Venture is left for future work.

\subsection{EXAMPLE 3: MODELING REAL-WORLD HUMAN MOTION} \label{sec:real-world-human-motion}
\begin{figure*}[ht]
    \centering
    \includegraphics[width=0.96\textwidth]{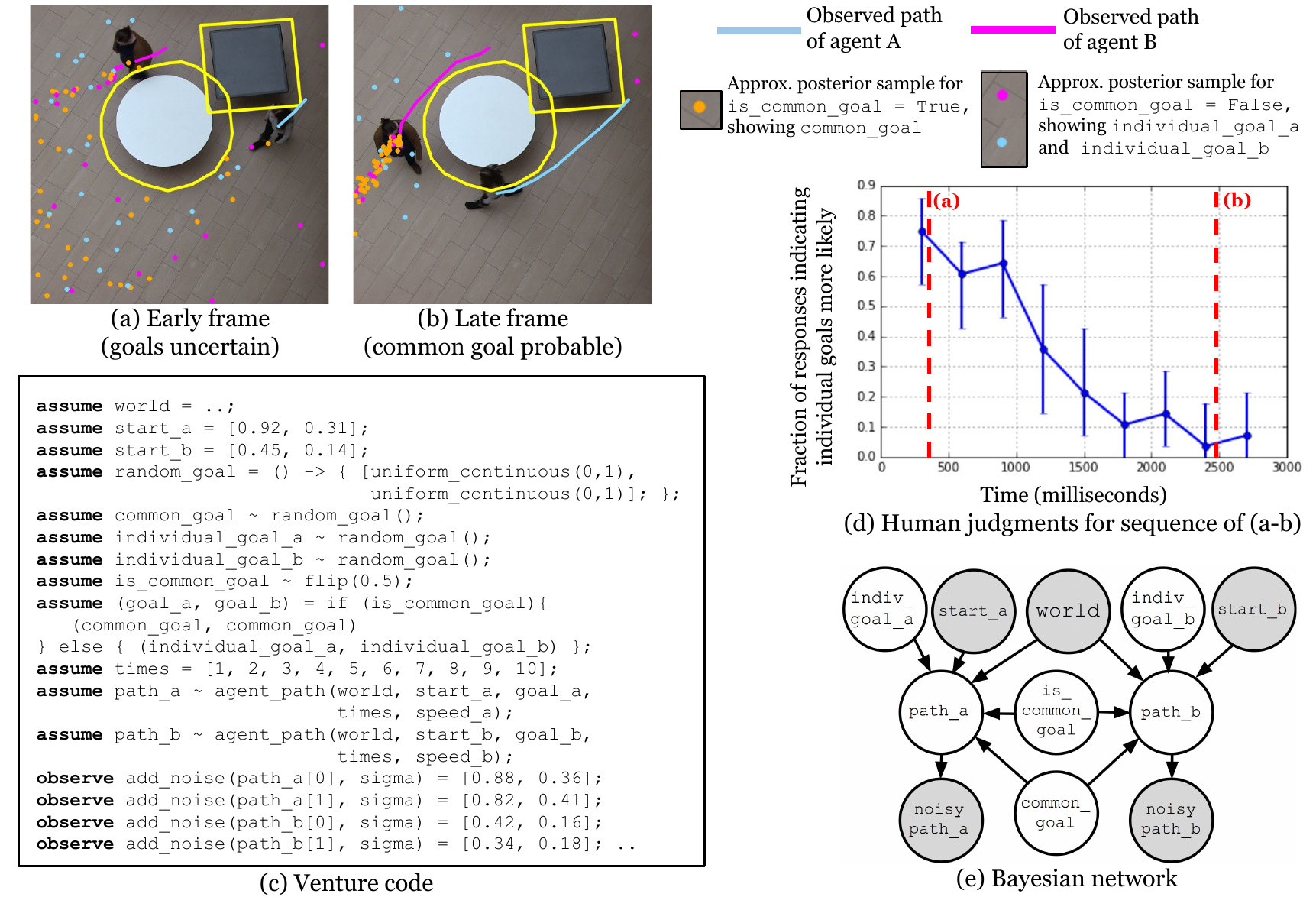}
    \caption{
Inferring whether or not two people are headed to the same destination. (c)
shows a Venture model of two people. (e) shows a Bayesian network
representation of the model with observed variables shaded. (a) and (b) show the map, two pairs of observed trajectories, and approximate
posterior samples obtained from Cascading Resimulation MH. Each sample with
{\small \texttt{is\_common\_goal=True}} is rendered as a single yellow circle. Each sample
with {\small \texttt{is\_common\_goal=False}} is rendered as two separate magenta and blue
circles. In (a) inference is uncertain about goal locations, and the estimated probability of {\small \texttt{is\_common\_goal=True}} is 0.63. In (b) inference gives a concentrated common goal region with estimated common-goal probability 0.82. Some probability mass is reserved for the people walking past one another to uncertain destinations. (d) shows judgments from 30 human responders of the likelihood over time that the individuals have different destinations, for the video sequence spanned by the frames in (a,b). The human judgments qualitatively agree with the automated inferences. }
    \label{fig:fig_real_data}
\end{figure*}

The Venture program of Figure~\ref{fig:fig_real_data}(c) defines a model with two
agents whose destinations may or may not be the same. The environment
({\small \texttt{world}}) and the start locations of the agents are known.
The {\small \texttt{is\_common\_goal}} flag determines whether the
agents share the same goal destination. The paths of both agents are modeled using
\textproc{agent-path}. The
corresponding Bayesian network is shown in Figure~\ref{fig:fig_real_data}(e).
We collected video of two collaborators walking in a scene containing tables, for two conditions---one in which the they meet at a common location, and one where they diverge. For the common-goal condition we constructed short and extended sequences of observed locations (Figure~\ref{fig:fig_real_data}(a) and (b)).
We used Cascading Resimulation MH for inference, initialized from the prior, with a joint prior proposal over all latent variables. We ran 60 chains of $200$
transitions each, and rendered the final states in Figures~\ref{fig:fig_real_data}(a-b). 
The speed for each individual was set to their average speed along the observed path.
The estimated probabilities of {\small \texttt{is\_common\_goal=True}} for the short and extended sequences are 0.63 and 0.82 respectively. 
This trend qualitatively matches human judgments, shown in Figure~\ref{fig:fig_real_data}(d) (the model was not calibrated to match human judgments).
See Appendix~\ref{sec:additional_experiments} for additional results.

\section{DISCUSSION}
This paper introduced a class of probabilistic programs for formulating goal inference as approximate inference in probabilistic generative models of goal-directed behavior. The technical contributions are (i) a probabilistic programming formulation that makes complex goal and map priors easy to specify; (ii) the use of randomized path planning algorithms as the backbone of generative models; and (iii) the introduction of Monte Carlo techniques that can handle the intractable likelihoods of these path planners. The experiments showed that it is possible for short probabilistic programs to make meaningful inferences about goal-directed behavior.

From the standpoint of robotics, autonomous driving, or reconnaissance, the examples in this paper are quite preliminary. More experiments are needed to explore the accuracy of approximate inference in these models, as well as the accuracy of the models themselves, especially on real-world problems. The probabilistic programming formulation makes it easy to explore variations of models, environments, and inference strategies.

The problem of inferring the mental states of autonomous agents is central to probabilistic artificial intelligence. It may also be a natural application for structured generative models and for probabilistic programming, but only if sufficiently fast and flexible inference schemes can be developed. We hope this paper helps to encourage the use of probabilistic programming for building intelligent software that can draw meaningful inferences about goal-directed behavior.

\subsubsection*{Acknowledgements}
The authors would like to thank Feras Saad for obtaining human judgment data, and Leslie Kaelbling, Erin Bartuska, and Feras Saad for helpful conversations.
Tree and car 3D models in figures are from \url{http://www.f-lohmueller.de/} \citep{models3d}.
This research was supported by DARPA (PPAML program, contract number FA8750-14-2-0004), IARPA (under research contract 2015-15061000003), the Office of Naval Research (under research contract N000141310333), the Army Research Office (under agreement number W911NF-13-1-0212), and gifts from Analog Devices and Google.  MCT is supported by the Department of Defense (DoD) through the National Defense Science \& Engineering Graduate Fellowship (NDSEG) Program.

\clearpage
\subsubsection*{References}
\renewcommand\refname{\vskip -6mm}
\bibliographystyle{plainnat}
\bibliography{references}

\clearpage
\appendix

\section{PLANNER DETAILS} \label{sec:planner-details}
We now describe details of the planner omitted from the main text, including the procedures \textproc{simplify-path}, \textproc{refine-path}, and \textproc{walk-path}, which are defined in Algorithm~\ref{alg:planner-details}. Paths $\mathbf{p}$ are represented as sequences of points, with lines connecting the points. The path $\mathbf{p}$ begins with start $s$ and ends with goal $g$. To be a valid path with respect to map $M$, no point in the path may lie within an obstacle (polygon) of $M$ (i.e. $M.\textproc{is-valid}(\mathbf{p}_i)$), and no line between two adjacent path points may intersect an obstacle of $M$ (i.e. $M.\textproc{clear-line}(\mathbf{p}_i, \mathbf{p}_{i+1})$).
\begin{algorithm}[H]
    \footnotesize
    \begin{algorithmic}[1]
    \Require $\left\{
        \begin{array}{l}
            \mbox{World map}\; M;\; \mbox{Start, goal}\; s,g \in [0,1]^2\\
            \mbox{Time points}\; \mathbf{t} \in \mathbf{R}_{+}^{T}\\
            \mbox{Refinement amount}\; N;\; \mbox{Restarts}\; R\\
            \mbox{Agent speed}\; v \in \mathbf{R}_{+}
        \end{array}
    \right.$
    \Procedure{simplify-path}{$M$, $\mathbf{p}$, $s$, $g$} 
        \State $\mathbf{p}'_1 \gets s$ \Comment{Initialize simplified path}
        \State $j \gets 2$
        \For{$i \gets 2$ to $\textproc{num-points}(\mathbf{p}) - 1$}
            \If{not $M.\textproc{clear-line}(\mathbf{p}_{i - 1}, \mathbf{p}_{i + 1})$}
                \State $\mathbf{p}'_j \gets \mathbf{p}_i$ \Comment{Point $\mathbf{p}_i$ is needed, keep it}
                \State $j \gets j + 1$
            \Else
                \State $\mathbf{pass}$ \Comment{Point $\mathbf{p}_i$ is not needed, skip it}
            \EndIf
        \EndFor
        \State $\mathbf{p}'_j \gets g$ \Comment{Add goal $g$ to simplified path}
        \State $\mathbf{return}\; \mathbf{p}'$
    \EndProcedure
    \Procedure{refine-path}{$M$, $s$, $g$, $\mathbf{p}$}
            \For{$i \gets 1$ to $N$}
            \State $d \gets \textproc{path-length}(\mathbf{p}, s, g)$
            \For{$l \gets 1$ to $L$} \Comment{{\small Iterate over $L$ path dims.}}
            \State $\epsilon \sim \mathcal{N}(0,\sigma^2)$ 
                \State ${\mathbf{p}}' \gets \mathbf{p} + \epsilon \cdot \mathbf{e}_l$ \Comment{{\small Change path dim. $l$}}
                \State $d' \gets \textproc{path-length}({\mathbf{p}}', s, g)$
                \If{{\small $d' < d \land M.\textproc{clear-path}({\mathbf{p}}', s, g)$}}
                    \State $(d, \mathbf{p}) \gets ({\mathbf{p}}',d')$ \Comment{Accept}
                \EndIf
            \EndFor
        \EndFor
        \State $\mathbf{return}\; \mathbf{p}$
    \EndProcedure
    \Procedure{walk-to}{$\mathbf{p}$, $t$, $v$}
        \State $d \gets 0.0$ \Comment{Path dist. from $s$ traveled so far}
        \State $d^* \gets t v$ \Comment{Desired path distance from $s$}
        \For{$j \gets 1$ to $\textproc{num-points}(\mathbf{p}) - 1$}
            \State $\delta \gets ||\mathbf{p}_{j} - \mathbf{p}_{j+1}||_2$ \Comment{{\small Dist. to next point}}
            \If{$d + \delta > d^*$}
                \State $e \gets d^* - d$
                \State $\mathbf{return}\; \frac{\delta - e}{\delta} \mathbf{p}_j + \left(1 - \frac{\delta - e}{\delta}\right) \mathbf{p}_{j+1}$
            \EndIf
            \State $d \gets d + \delta$
        \EndFor
        \State $\mathbf{return}\; g$ \Comment{Once reached goal, stay forever}
    \EndProcedure

    \Procedure{walk-path}{$\mathbf{p}$, $\mathbf{t}$, $v$}
        \For{$i \gets 1$ to $T$}
            \State $\mathbf{z}_i \gets \textproc{walk-to}(\mathbf{p}, t_i, v)$
        \EndFor
        \State $\mathbf{return}\; \mathbf{z}$
    \EndProcedure
    \end{algorithmic}
    \caption{Additional details of the \textproc{agent-path} model of goal-directed behavior.}
    \label{alg:planner-details}
\end{algorithm}

\section{ADDITIONAL EXPERIMENTS} \label{sec:additional_experiments}

\subsection{JOINTLY INFERRING THE BELIEF AND GOAL OF AN AGENT}

The Venture program of Figure~\ref{fig:joint-map-goal-inference}(a) defines a model
in which the belief of an agent about its environment, upon which the agent's motion plan depends, is uncertain.
The environment contains two, static 
objects (\texttt{known\_objects}): a tree and a central divider wall that divides the
$[0,1]\times[0,1]$ square into a left and right side. There are passageways
between the left and right side that go above and below the divider. However,
the agent has knowledge of (or belief in) an additional obstacle
wall (\texttt{obstacle}), and the agent plans their path to the destination
(\texttt{goal}) taking this additional obstacle into account.
Figure~\ref{fig:joint-map-goal-inference}(a) also shows a Bayesian network representation of this model.
We seek to infer both the agent's goal and the agent's beliefs about the
location, orientation, and size of the obstacle.

We used Cascading Resimulation Metropolis-Hastings (Algorithm~\ref{alg:cascading-mh}) with a
single repeated transition operator based on an independent joint proposal to
\texttt{goal} ($\mbox{Uniform}([0,1]^2)$) and to the unknown
parameters of \texttt{obstacle} (start post location, orientation, and length,
proposed from the prior). We initialized from the prior.
Parameters of the planner \textproc{agent-path} were $R = 10$, $N = 1000$, $\epsilon = 0.01$, $v =0.5$, $J = 10000$, $S = 2000$.
We ran several independent Markov chains of $1000$ iterations each, on a synthetic dataset in which the agent takes a path from the right to the left of the map by going below the divider. The final state of four such chains are visualized in Figure~\ref{fig:joint-map-goal-inference}(b).
For this dataset, the goal destination of the agent is revealed with certainty because the agent reaches and stops in the upper left corner.
The obstacle inferences indicate that agent believes the upper route to its goal is blocked, because otherwise the agent would have taken the shorter, upper route, to its goal.
However, the specific
details of how the obstacle blocks the upper passageway remain uncertain.

\subsection{GOAL INFERENCE IN A DRIVING SCENARIO}
Figure~\ref{fig:fig3} shows an application of the multi-agent common-goal model of
Figure~\ref{fig:fig_real_data} to a driving scenario. We show 60 independent replicates of 3000 iterations
of Cascading Resimulation Metropolis-Hastings each. The results illustrate that
this model can be used with varied environments.

\subsection{REAL-WORLD HUMAN MOTION, ALTERNATE SEQUENCE}
\begin{figure}[h]
    \centering
    \includegraphics[width=0.48\textwidth]{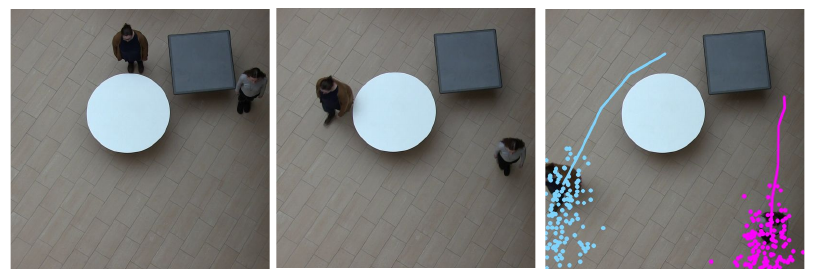}
    \caption{Inferring whether or not two people are headed to the same destination, as in Figure~\ref{fig:fig_real_data}, but for a different sequence of observed locations. The final frame shows approximate
posterior inference samples obtained from cascading
resimulation Metropolis-Hastings. Inference gives low probability of a common goal 
for this sequence (there were no \texttt{is\_common\_goal = True} samples).}
    \label{fig:extra_real}
\end{figure}

\begin{figure*}[h]
    \centering
    \includegraphics[width=1.00\textwidth]{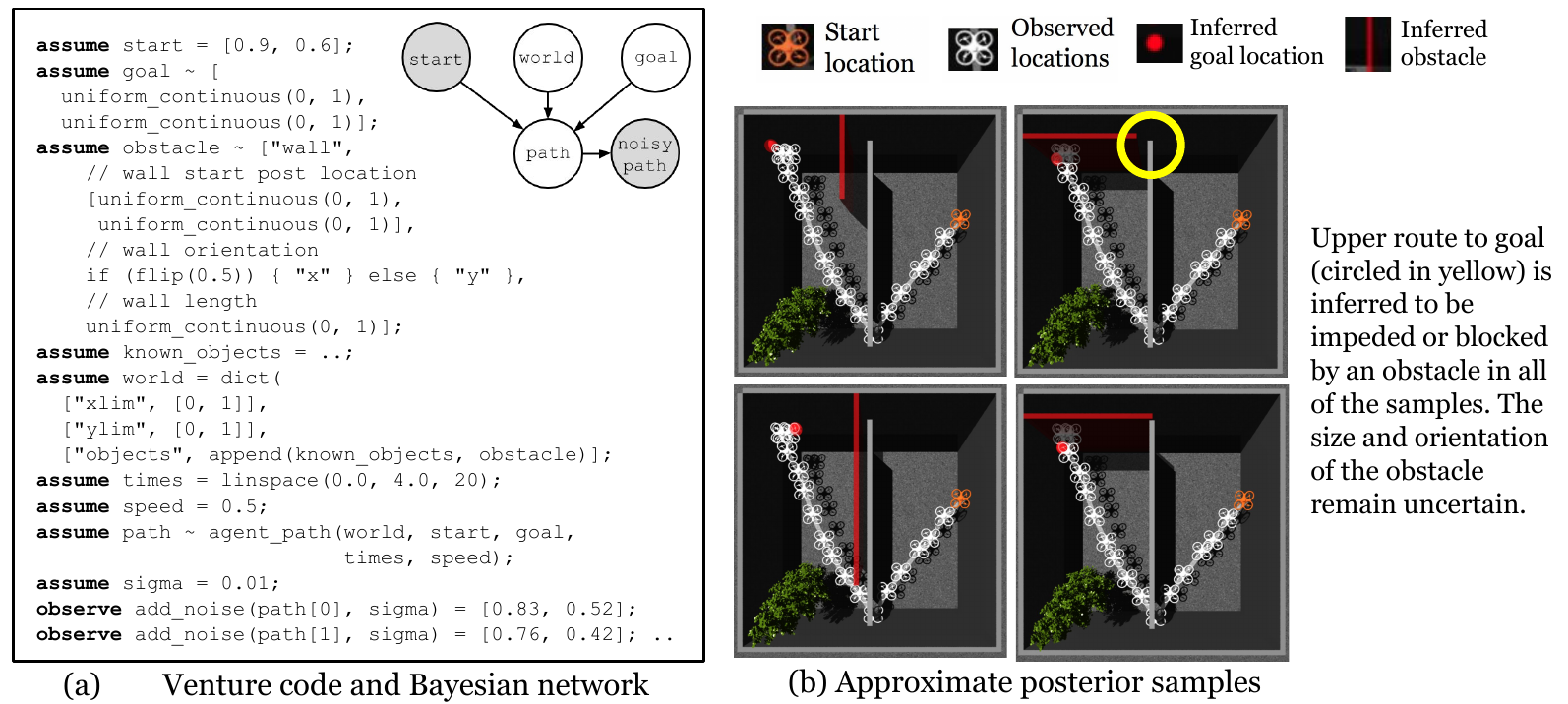}
    \caption{
Inferring the belief of an agent about the location and shape of an
obstacle in its environment, from
observations of the agent's motion. (a) shows a Venture model of the
agent's belief, goal, and resulting motion and a Bayesian network
representation of the model with observed variables shaded. (b) shows
approximate posterior samples of \texttt{goal} and \texttt{obstacle} obtained
with Cascading Resimulation Metropolis-Hastings, for a data set in which the goal is disambiguated to lie in the upper left corner.  
The obstacle samples in (b) indicate that inference
in the model concluded that the
agent believes that there is an obstacle blocking the upper route to its goal. Otherwise, the agent would have taken the shorter, upper route.
}
    \label{fig:joint-map-goal-inference}
\end{figure*}

\begin{figure*}[h]
    \centering
    \includegraphics[width=1.00\textwidth]{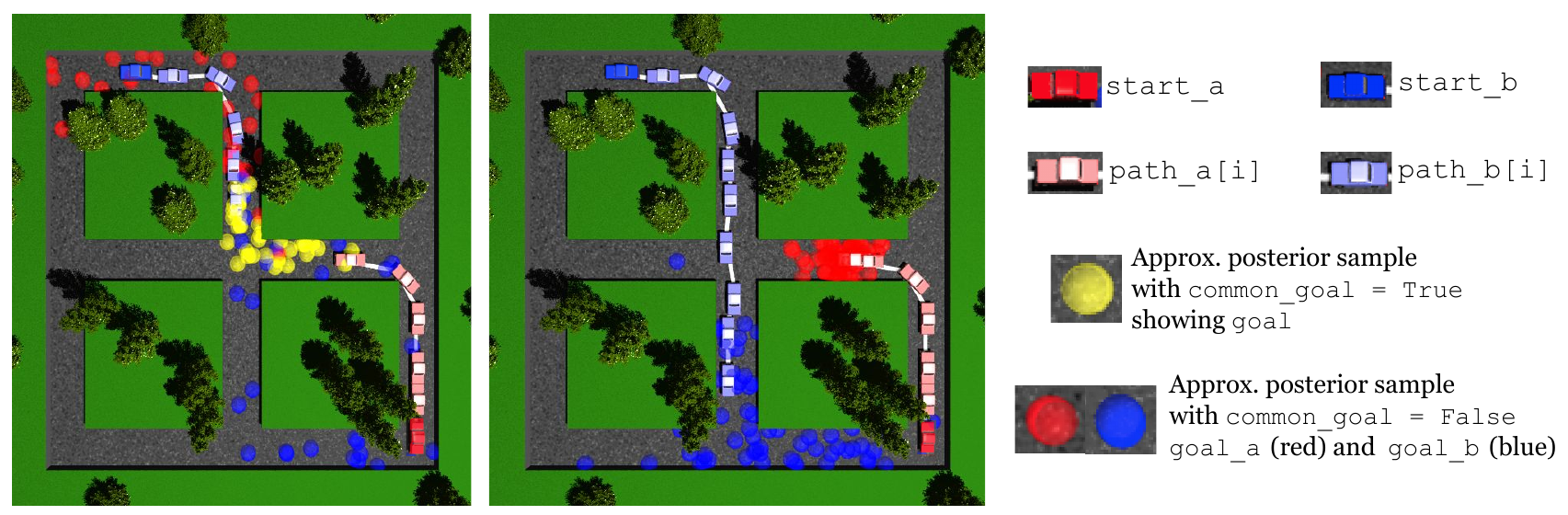}
    \caption{
A synthetic application of the common-goal inference problem from
Figure~\ref{fig:fig_real_data} to a different scenario inspired by autonomous driving.
Above shows approximate posterior inference samples obtained with independent
runs of Cascading Resimulation MH. Each sample with \texttt{is\_common\_goal =
True} is rendered as a single yellow sphere. Each sample with
\texttt{is\_common\_goal = False} is rendered as two blue and red spheres.
Left: Inference indicates an approximate 0.5 probability that the cars are
both headed for the center of the map. Right: The red car has stopped,
revealing its goal, and the blue car continued, indicating that the
two cars do not share the same destination.}
    \label{fig:fig3}
\end{figure*}

We extended the experiment described in Section~\ref{sec:real-world-human-motion} and shown in
Figure~\ref{fig:fig_real_data} by running Cascading Resimulation
Metropolis-Hastings on an alternate sequence of observed person locations in
which the individals diverge to separate individual goal destination. The
inferences, shown in Figure~\ref{fig:extra_real}, confirm the expectations, with all
samples indicating \texttt{is\_common\_goal = False}. Samples were obtained from
the final state of $120$ independent Markov chains, with initialization from
the prior, followed by $1200$ iterations of Cascading Resimulation Metropolis-Hastings.

\subsection{INFERENCE WITH WAYPOINT PLANNER} \label{sec:waypoint-additional}
\begin{figure}[H]
    \centering
    \includegraphics[width=0.48\textwidth]{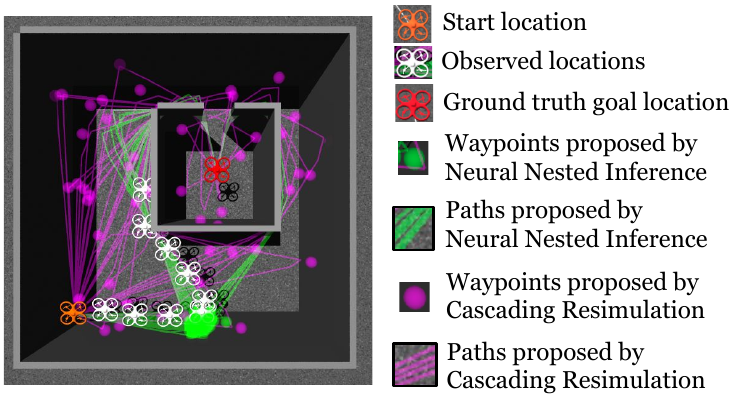}
    \caption{
        Proposed waypoints and paths produced by Nested Inference MH with a neural nested inference algorithm, and Cascading Resimulation MH, when evaluating the MH acceptance ratio for a proposed goal $g'$ in the center of the enclosure that is the ground truth goal. Because the neural nested inference algorithm generates reasonable proposed waypoints, the proposed paths have a high probability of being consistent with the observed data. 
Because Cascading Resimulation proposes the waypoint and path from the prior for each proposed goal $g'$, the paths proposed are unlikely to be consistent with the observations, resulting in a high MH rejection rate, even when the proposed goal $g'$ is the ground truth goal, as is the case here.
        }
    \label{fig:fig_waypoint_detail}
\end{figure}

\vspace{-1mm}
Figure~\ref{fig:fig_waypoint_detail} compares waypoints and paths proposed by Nested Inference MH with a neural nested inference algorithm with waypoints and paths proposed by Cascading Resimulation MH on an illustrative example data set. The poor quality of the prior as the proposal, as used by Cascading Resimulation, results in unecessary rejections, and slow convergence. The neural network proposes waypoints near the bend in the path.

\begin{figure*}[h]
    \centering
    \includegraphics[width=1.00\textwidth]{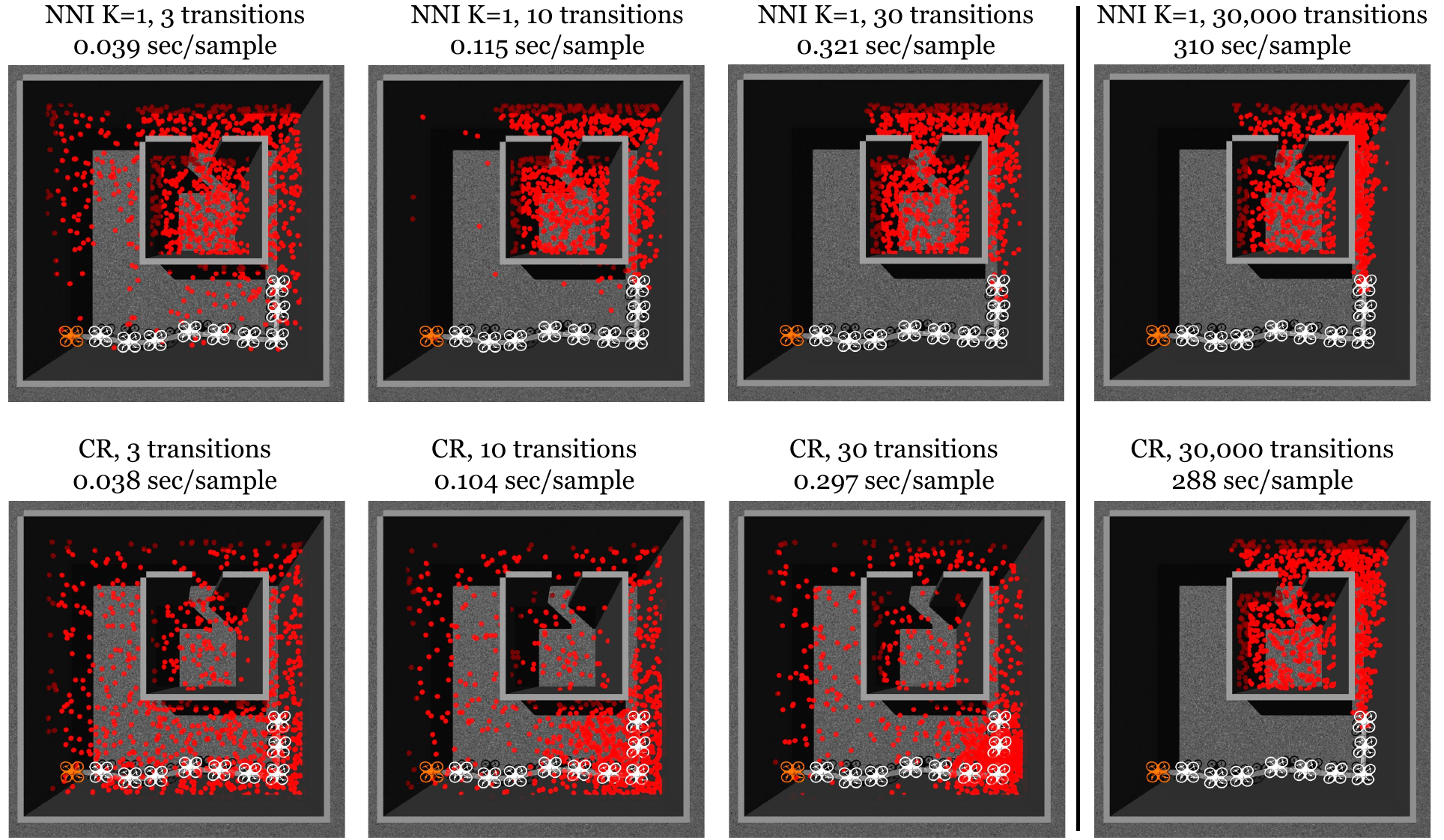}
    \caption{Additional comparisons of approximate posterior goal inferences using neural Nested Inference Metropolis-Hastings (NNI, top row) and Cascading Resimulation Metropolis-Hastings (CR, bottom row). The drone starts in the lower left corner (orange). Observations of the drone's location are shown in white. Red dots are independent approximate posterior samples of the drone's goal. The neural Nested Inference MH strategy converges to a qualitatively correct distribution within 100-300ms (indicating real-time performance), whereas Cascading Resimulation MH requires 10-30 seconds to produce similarly accurate inferences (also see Figure~\ref{fig:nested-combined}(d)).}
    \label{fig:fig_waypoint_additional}
\end{figure*}

The KL divergence estimates of Figure~\ref{fig:nested-combined}(d) were obtained by binning 960 independent reference samples (30,000 transitions of Cascading Resimulation MH, initialized from the prior) and binning 960 independent approximate inference samples for each inference algorithm evaluated. The world unit square was binned into 25 squares (5-by-5), and a discrete distribution was estimated for each sampler by counting the number of samples falling into each bin, adding a pseudocount of $0.1$ to each bin, and normalizing. The KL divergence from the resulting reference sampler histogram was computed to each resulting approximate inference algorithm histogram. For each inference strategy (Cascading Resimulation MH, Neural Nested Inference MH with $K=1$, Resimulation Nested Inference MH with $K=2$ and Resimulation Nested Inference MH with $K=10$), the number of MH transitions was varied over several orders of magnitude, and the final state in each chain was recorded, to obtain samples for each inference algorithm evaluated.
The number of MH transitions used to obtain samples shown in Figures~\ref{fig:nested-combined}(a,b,c) are 10, 1, and 10, respectively. Figure~\ref{fig:fig_waypoint_additional} shows additional samples comparing Nested Inference Metropolis-Hastings with a neural nested inference algorithm with Cascading Resimulation Metropolis-Hastings. 

\clearpage
\section{NESTED INFERENCE DERIVATIONS} \label{sec:nested-inference-derivations}
The variance of the likelihood estimate with $K = 1$ is:
\begin{align*}
&\mbox{Var}_{u \sim q(\cdot; x, z)} \left[ \frac{p_t(u, z; x)}{q_t(u; x, z)} \right]\\
&\qquad= p_t(x;z)^2 \cdot \int \left( \frac{p_t(u| z; x)}{q_t(u; x, z)} - 1 \right)^2 q(u; x, z) du\\
&\qquad\propto \int \left( \frac{p_t(u| z; x)}{q_t(u; x, z)} - 1 \right)^2 q(u; x, z) du\\
&\qquad= \mbox{D}_{\chi^2}(p_t(u|z;x) || q_t(u; x, z))
\end{align*}
The bias of the log likelihood estimate with $K = 1$ is:
\begin{align*}
    & \mbox{E}_{u \sim q_t(\cdot; x, z)} \left[ \log \frac{p_t(u, z; x)}{q_t(u;x,z)} \right] - \log p_t(z;x)\\
    &\qquad = \mbox{E}_{u \sim q_t(\cdot; x, z)} \left[ \log \frac{p_t(u| z; x)}{q_t(u;x,z)} \right]\\
    &\qquad = -\mbox{D}_{KL}( q_t(u; x, z) || p_t(u | z; x) )
\end{align*}
Algorithm~\ref{alg:nested-inference-mh} can be intuitively understood as an approximation to a single-site MH update where a single value $z_i$ is being updated with proposal $m(z_i'; z)$ and target density $p(z_i | z_{I \setminus \{i\}})$, and where estimates of likelihoods are used in place of actual likelihoods when computing the MH acceptance ratio.
However, Algorithm~\ref{alg:nested-inference-mh} is theoretically justified by recognizing that it is a standard joint MH transition on an extended state space that consists of $z_i$ (the value of the proposed-to random choice), $u_{i,k}$ for $k=1\ldots K$ (a set of $K$ traces for the proposed-to random choice), and $u_{j,k}$ for $j \in c_G(i)$ and $k=1\ldots K$ (a set of $K$ traces for each of the children of the proposed-to random choice). The extended target density is:
\begin{equation}
\resizebox{\hsize}{!}{
    $\displaystyle p(z_i | z_{I \setminus \{i\}}) \prod_{j \in \{i\} \cup c_G(i)} \frac{1}{K} \sum_{k=1}^K p_{t_j}(u_{j,k} | z_j; x_j) \prod_{\substack{r=1\\r \ne k}}^K q_{t_j}(u_{j,r}; x_j,z_j)$
}
\end{equation}
Note that the marginal target density of $z_i$ is the original target density $p(z_i | z_{I \setminus \{i\}})$,
which is proportional to $p_{t_i}(z_i; x_i) \prod_{j \in c_G(i)} p_{t_j}(z_j; x_j)$.
Substituting $p_{t_i}(z_i; x_i) \prod_{j \in c_G(i)} p_{t_j}(z_j; x_j)$ for $p(z_i | z_{I \setminus \{i\}})$ in the extended target density expression and simplifying gives the following unnormalized extended target density:
\begin{align}
\prod_{j \in \{i\} \cup c_G(i)} \frac{1}{K} \sum_{k=1}^K p_{t_j}(u_{j,k}, z_j; x_j) \prod_{\substack{r=1\\r \ne k}}^K q_{t_j}(u_{j,r}; x_j,z_j)
\end{align}
The extended proposal density is:
\begin{equation}
    m(z_i'; z) \prod_{j \in \{i\} \cup c_G(i)} \prod_{k=1}^K q_{t_j}(u_{j,k}'; x_j, z_j)
\end{equation}

The ratio of the unnormalized extended target density over the extended proposal density, for proposed values $z_i'$ and $u_{j,k}'$ for all $j \in \{i\} \cup c_G(i)$ and $k=1\ldots K$, with all other $z_{I \setminus \{i\} \cup c_G(i)}' = z_{I \setminus \{i\} \cup c_G(i)}$ is:
\begin{align}
\frac{1}{m(z_i';z)} \prod_{j \in \{i\}\cup c_G(i)} \frac{1}{K} \sum_{k=1}^K \frac{p_{t_j}(u_{j,k}', z_j'; x_j')}{q_{t_j}(u_{j,k}';x_j',z_j')}
\end{align}
Note that each factor $\frac{1}{K} \sum_{k=1}^K \frac{p_{t_j}(u_{j,k}, z_j; x_j)}{q_{t_j}(u_{j,k};x_j,z_j)}$ within this ratio takes the form of a nested inference likelihood estimate. Composing the full MH acceptance ratio for the extended target and proposal densities gives the acceptance ratio used in Algorithm~\ref{alg:nested-inference-mh}. 
Note that Algorithm~\ref{alg:nested-inference-mh} samples the proposed joint state $z_i$,  $u_{i,k}$ for $k=1\ldots K$, and $u_{j,k}$ for $j \in c_G(i)$ and $k=1\ldots K$ precisely according to the extended proposal density.
Finally, note that the nested-inference likelihood estimates $\ell$ for accepted proposals are retained between updates in Algorithm~\ref{alg:nested-inference-mh}. These estimate values serve as summaries of the previous iterates for the traces $u_{i,k}$ and $u_{j,k}$. Although the transition is an MH transition on the extended space including the traces, the previous estimates $\ell$ are sufficient for evaluating the extended MH acceptance ratio and retaining the previous trace iterates themselves is not necessary.

\end{document}